\pdfoutput=1

\documentclass[11pt]{article}

\usepackage[]{acl}

\usepackage{times}
\usepackage{latexsym}

\usepackage[T1]{fontenc}

\usepackage[utf8]{inputenc}

\usepackage{microtype}

\usepackage{inconsolata}
\usepackage{url}
\usepackage{amsmath}
\usepackage{graphicx}
\title{``Breaking the Silence'' Detecting and Mitigating Gendered Abuse in Hindi, Tamil, and Indian English Online Spaces}

\author{Advaitha Vetagiri$^{1}$, Gyandeep Kalita$^{1}$, Eisha Halder$^{1}$, Chetna Taparia$^{1}$, Partha Pakray$^{1}$, Riyanka Manna$^{2}$ \\
  $^{1}$National Institute of Technology Silchar, 
  Silchar,
  Assam,
  India. \\
  $^{2}$Computer Science \& Engineering, School of Computing, Amrita Vishwa Vidyapeetham\\
  Amaravati Campus,
  Andhra Pradesh,
  India. \\
  \texttt{$^{1}$advaitha21\_rs@cse.nits.ac.in, $^{1}$gyandeepkalita1@gmail.com, $^{1}$eishashalder@gmail.com,}\\
  \texttt{$^{1}$chetna.taparia@gmail.com, $^{1}$partha@cse.nits.ac.in,}
  \texttt{$^{2}$m\_riyanka@av.amrita.edu}
}

\begin{document}
\maketitle
\begin{abstract}
Online gender-based harassment is a widespread issue limiting the free expression and participation of women and marginalized genders in digital spaces. Detecting such abusive content can enable platforms to curb this menace. We participated in the ``\textit{Gendered Abuse Detection in Indic Languages}'' shared task at ICON2023 that provided datasets of annotated Twitter posts in English, Hindi and Tamil for building classifiers to identify gendered abuse. Our team \textbf{CNLP-NITS-PP} developed an ensemble approach combining CNN and BiLSTM networks that can effectively model semantic and sequential patterns in textual data. The CNN captures localized features indicative of abusive language through its convolution filters applied on embedded input text. To determine context-based offensiveness, the BiLSTM analyzes this sequence for dependencies among words and phrases. Multiple variations were trained using FastText and GloVe word embeddings for each language dataset comprising over 7,600 crowdsourced annotations across labels for explicit abuse, targeted minority attacks and general offences. The validation scores showed strong performance across f1-measures, especially for English 0.84. Our experiments reveal how customizing embeddings and model hyperparameters can improve detection capability. The proposed architecture \textbf{ranked 1st} in the competition, proving its ability to handle real-world noisy text with code-switching. This technique has a promising scope as platforms aim to combat cyber harassment facing Indic language internet users. Our Code is at \url{https://github.com/advaithavetagiri/CNLP-NITS-PP}
\end{abstract}

\section{Introduction}
Sexism exists in online communication today just as it does offline. Sexism continues because modern online communication is just as vital a part of current society as anything else, and there are benefits to it, just as there are disadvantages. Online chat, such as sexist language, leads to harassment, cyberspace bullying, and gendered discourse. Indicates this \cite{whiley2023contributions}, as argued by \cite{hoskin2023femme} \cite{felmlee2023debating}, both males and females feel the impacts of toxic masculinity Sexism is an acknowledged problem in internet-based discussions. However, it is still difficult to distinguish it.

As observed by \citep{kural2022attachment}, this has serious ramifications affecting self-esteem, anxiety, and feelings of insecurity among the targets. As stated by \citep{feigt2022impact}, these adverse effects could have a prolonged impact on one’s life in terms of deteriorating mental health and relationships. Finally, using sexist language further contributes towards inequality among genders because it continues to propagate biased messages, which is explained more by \citep{barreto2023benevolent}.

One crucial area of research involves automatically detecting sexist language in text. Automated algorithms can help decrease the occurrence of sexism and make the online environment more inclusive \citep{vetagiri4733628multilate}; one case is when somebody publishes sexist words. However, it presents various challenges in developing autonomous systems capable of consistently identifying sexist language in texts, such as interpreting the context of language use and cultural nuances of language \citep{van2015critical}. These problems become more complex because of Internet communication, where people often use abbreviations and shorthand that may not be easily understood.

Online gender-based violence is taking the upper hand among other challenges, and it is an addition of social and economic weaknesses, especially for the people in Indic languages. Such abuse scares people away from virtual space, which subsequently affects individuals’ political or economic prospects. It manifests itself very seriously and can be tragic, even leading to death. In light of this, it is essential to develop an automatic mechanism that identifies gender insults as they occur in interactions taking place in Indic languages. However, the development of these approaches is hampered by a significant shortage of Indic language-dependent datasets.

Under ICON 2023\footnote{\url{http://icon2023.unigoa.ac.in/}}, led by Tattled Civic Tech\footnote{\url{https://tattle.co.in/}}, we explore a relevant topic on how to counter online gender-based violence in Indic languages among participants in a shared task. This challenge multiplies present social and economic insecurities such that they could drive people out of social media platforms, disrupt their chances of politically or economically surviving, and sometimes result in deaths. To develop fast techniques for identifying gender-based violence, we acknowledge the lack of Indic language data sets essential for advancing this field of study. Our approach has mainly centred on assisting in a solution, paying specific attention to Hindi, Tamil, and Indian English. This allows us to access the new, well-annotated data from 18 activists and researchers based on their observations and experience with gender violence. We rely on this dataset \cite{arora2023uli} incorporating 7638 posts in English, 7714 in Hindi, and 7914 in Tamil for the basis of our involvement in the shared task.

\section{Related Work}

Research on gender-based online harassment has gained prominence due to its pervasive and damaging effects on individuals. The early attempts include machine learning techniques for detecting gender-based online harassment. Various models, e.g., Support Vector Machines (SVMs) \cite{ghosal2023inculcating}, Random Forests \cite{das2023performance}, have been used to classify abusive and non-abusive content. These methods have potential, but there is a great difficulty in obtaining a high accuracy that fits different language structures and cultural circumstances.

The modern day developments in deep learning techniques have made it possible to develop better approaches for detecting offensive languages on e-platforms today. CNNs \cite{quoc2023vietnamese} and RNNs \cite{jahan2023systematic} efficiently pick up complicated attributes and context details. The models mentioned above that are trained with high data set sizes can identify slight differences in gender discriminative abuse.

The ensemble models especially, which incorporate Convolutional Neural Networks (CNNs) and Bi-directional Long Short-Term Memory networks (BiLSTMs) \cite{vetagiri-etal-2023-cnlp}, are one of the most effective techniques in dealing with the complex nature of the problem of gender-based online The following part discusses the benefits and effectiveness of ensembled architecture and specifically the CNN-BiLSTM blend. Some ensemble models, such as the CNN-BiLSTM \cite{vetagiri2023leveraging}, combine the advantages of CNNS and bi-LSTMs for capturing spatial and temporal dependencies inherent in textual data. CNNs are good at extracting local features and patterns, whereas biLSTMs capture long-range dependencies, thus providing complimentary components for unravelling the complexities of abusive language.

The transferability of models trained in one language for use with another language has become an area of interest. Preliminary studies \cite{priyadarshini2023transfer} have focused on training models in English for abusive language development before being fine-tuned with Indic languages. The second method uses abundant English-language resources and overcomes the lack of labelled Indian data.

Researchers have studied multimodal approaches that consider different modes of abusive content, such as text, pictures, and videos. Deep learning-based fusion of text and image features \cite{chhabra2023literature} appears promising at reflecting diverse Gender-based violence (GBV) expression in different genres.

An important issue in implementing machine learning models for monitoring harassment based on gender is the consideration of cultural and language features. Some studies \cite{ghosal2023inculcating, das2023performance} stress that when designing models for indigenous languages like Hindi, we should be aware of local linguistic peculiarities and cultural specificity.

\section{Dataset}

The dataset creation process began with a focus on three Indian languages: ‘Indian’ English is the name used for the language that has become distinguishable as it involves transliteration and code-switching; Indian English, Hindi, and Tamil. A list of slurs and offensive phrases was crowd-sourced from activists and researchers to create a widely varied dataset. To this end, various lists of accounts commonly associated with Hate Online and Hate Speech offenders were also established. The researcher used Twitter data spanning between 2018 and 2021. Such criteria included slurs that are crowdsourced, tweets by perpetrators, and responses to influential women. 

The user handles were appropriately anonymized, and Python’s Twint library was used to obtain datasets. Subsequently, the stratified pooling technique selected 8000 instances out of the 1.3 million unmarked posts pool. The unsupervised dataset was labelled using democratic co-training of several models, whose parameters were pre-trained on public source datasets for similar tasks. Posts were then selected for the final data set by calculating average confidence scores using the mixture-of-experts (Moe) model based on ten classes having different toxicity score levels.

A carefully collected dataset \cite{arora2023uli} for the Hindi, Tamil, and Indian English shared tasks on detecting gender-based cyber violence. It consists of 7638 posts in English, 7714 in Hindi, and 7914 in Tamil. Eighteen activists carefully annotated each post, and researchers had first-hand or expert knowledge of gender studies. 

All languages and labels are compiled into a single dataset for convenience purposes. The dataset consists of three labels; each tweet is shown three times as it reflects the annotation of the different labels. The dataset columns include:
\begin{itemize}
    \item Label 1 (question\_1): Determines if the post amounts to gendered abuse, especially when targeting people who are not from gender or sexually marginalised groups.
    \item Label 2 (question\_2): Evaluate the extent to which it would qualify as gendered abuse to a person of minority gender and sexual orientation.
    \item Label 3 (question\_3): Indicates whether the post is overtly hostile.
\end{itemize}

\begin{table}
    \centering
    \begin{tabular}{cc}\hline
       Language &  Count\\
       \hline
       English  & 6531\\
       Hindi  & 6197\\
        Tamil & 6778\\
        \hline
    \end{tabular}
    \caption{The training dataset size per language}
    \label{tab:tabel1}
\end{table}

\begin{table}
    \centering
    \begin{tabular}{cc}\hline
       Language &  Count\\
       \hline
       English  & 7638\\
       Hindi  & 7714\\
        Tamil & 7914\\
        \hline
    \end{tabular}
    \caption{The total dataset size per language}
    \label{tab:tabel2}
\end{table}
Each post is annotated by the assigned annotators with values such as ``\textit{1}'' to indicate agreement with the label, ``\textit{0}'' to denote disagreement, ``\textit{NL}'' for posts that were assigned but not annotated, and ``\textit{NaN}'' for posts that were not assigned to annotators.

As shown in the Tables \ref{tab:tabel1} \& \ref{tab:tabel2}, this makes it easy as all languages with associated labels are collected in the same file. This means that each tweet is denoted as one of the three labels three times in the dataset. The dataset columns include:
\begin{itemize}
    \item id: Seq no—a serial number for each row.
    \item text: The content of the tweet.
    \item language: The tweet’s language (English, Hindi, or Tamil).
    \item key: Label identifier (question 1, question 2, or question 3).
    \item en\_a1 ... en\_a6: Annotations by English annotators; presenting the assigned values.
    \item hi\_a1 ... hi\_a5: Annotations per tweet for Hindu annotators and columns.
    \item ta\_a1 ... ta\_a6: Values annotators assign to columns for Tamil annotators.
\end{itemize}

The dataset \footnote{\url{https://github.com/tattle-made/uli_dataset}} is comprehensive in its structure. This will be an opportunity for the researchers and participants of the shared task to explore, analyze and formulate effective models of gender-based abuse detections in Indic languages.

\section{Tasks Description}

Indeed, we are happy to participate in the Shared Task, named “\textit{Gendered abuse detection in indic languages}” within ICON 23. This objective project intends to address the ever-escalating problem of cyber sexual harassment, whose impacts run deep in society and the economy as well. Our engagement in this shared task involves addressing three subtasks:
\begin{itemize}
    \item Build a Classifier for Gendered Abuse (Label 1): The task is to develop a classifier from the available dataset concerning label 1, gender-based abuse. Using carefully made eighteen activist’s and researcher’s annotations, we aim to create a robust system of detecting gender-based cyber violence in online communication.
    \item Transfer Learning for Gendered Abuse Detection (Label 1): We intend to use transfer learning from other open datasets concerning hate speech and toxic text recognition in the Indic languages for this subtask. We, therefore, improve our gendered abuse detection model by seeking to incorporate external knowledge and patterns to strengthen the overall effectiveness of the entire detection device.
    \item Multi-Task Classifier for Gendered Abuse and Explicit Language (Labels 1 and 3): We also take part in constructing a multitask classifier that will simultaneously estimate gendered abuse (label 1) and explicit language (label 3). In turn, this subtask is per our undertaking to consider gender-based cyberbullying as well as harsh/explicit vocabulary.
\end{itemize}

It is worth noting that we are involved in this shared endeavour to demonstrate our willingness to contribute towards creating methodologies of computerized detection of gender discrimination on the internet. It excites us to work together, learn, and take tangible steps towards developing a secure Internet environment.

\section{Methodology}

\subsection{System Overview}
After thorough experimentation with various neural network architectures, pretrained Large Language models, and classical machine learning models, we settled on an ensemble approach, developing a sophisticated model built upon a Convolutional Neural Network (CNN) and Bidirectional Long Short-Term Memory (BiLSTM) architecture, for all three Tasks.  

\begin{figure}[!htp]
    \centering
    \includegraphics[width=0.9\linewidth]{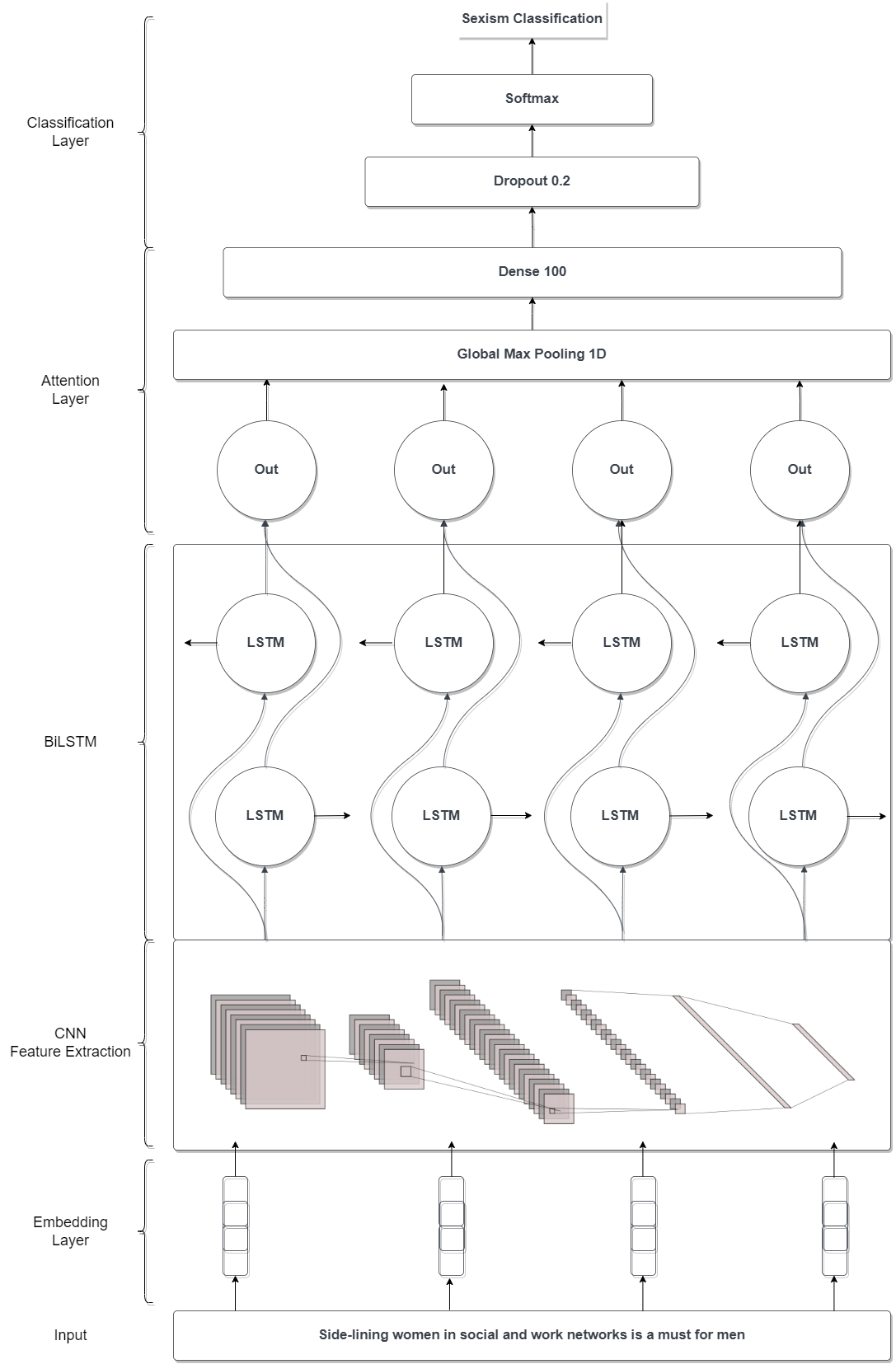}
    \caption{CNN-BiLSTM Architecture for Sexism Classification}
    \label{fig:CNN_Arc}
\end{figure}

The model implements a combination of CNN layers for capturing localized textual features indicative of abusive language in the input text and Bidirectional LSTM Layers strategically employed for comprehending the intricate, long-term, complex sequential dependencies within the text data. This synergistic integration enables the model to navigate and discern nuanced patterns, facilitating robust performance in classifying abusive and sexist language across diverse linguistic contexts.

For the initial input layers, the model uses pre-trained GloVe and FastText embeddings  \cite{kumar-etal-2020-passage} for the respective languages, representing the words as 300-dimensional dense vectors, with the sequence length capped at 100 words, held as non-trainable parameters. These word embeddings map each token to a vector of real numbers aiming to quantify and categorize the semantic similarities between the linguistic terms based on their distributional properties in a large corpus using machine learning or related dimensional reduction techniques. This was followed by a SpatialDropout1D layer, a dropout variant that selectively drops entire 1D feature maps during training to combat overfitting. 

\begin{figure}[!htp]
    \centering
    \includegraphics[width=0.9\linewidth]{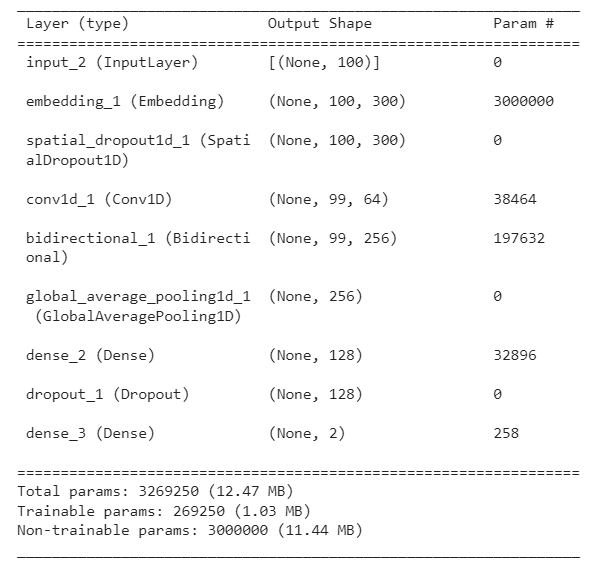}
    \caption{CNN-BiLSTM Model Summary}
    \label{fig:Model_Summary}
\end{figure}

For the CNN-BiLSTM Layers as shown in Figure \ref{fig:CNN_Arc}, a one-dimensional convolution layer employing 64 filters and a kernel size of 2 was employed to capture localized textual patterns with refined granularity. The output for this layer was then passed through a Bidirectional LSTM layer featuring 128 units and a return sequence setting, coupled with a dropout of 0.1 and recurrent dropout of 0.1, to process textual inputs bidirectionally. This facilitated comprehensive analysis in both forward and reverse directions. A dense layer with 128 neurons and Global Average Pooling is applied for dimensionality reduction and holistic sequence information aggregation. At last, a dropout layer is employed, whose output is passed through a dense layer with a softmax activation function to generate the classification Model. A comprehensive summary of the model including all the layers and trainable/non-trainable parameters has been portrayed in Figure \ref{fig:Model_Summary}

\subsection{Experimental Setup}
We trained different models for each of the three languages, viz. English, Hindi, and Tamil, for each of the three subtasks, using the CNN-BiLSTM architecture mentioned in the previous subsection, the details of which are discussed in the following sub-subsections.

\subsubsection{Tasks 1 \& 3}
As mentioned in the previous sections, Task 1 required us to develop a classifier from the
available dataset concerning label 1 and gender-based abuse. Task 3 required us to construct a multitask classifier to estimate gendered abuse (label 1) and explicit language (label 3) simultaneously. For both these tasks, we used the provided labeled dataset for Label 1 for Task 1 and Label 1 \& 3 for Task 3. 

We first calculated the final label for each sentence for datasets of both Labels for all languages. This was achieved by considering the majority occurrence of 0 or 1 among all annotators. The case of equal occurrences of 0s and 1s, the final label was considered as 1. The datasets were then preprocessed to remove stopwords, symbols, tags, and emojis, which we believed would result in a better generalization of the models, leading to superior performance compared to the raw datasets. The datasets were then divided into an 80/20 train-test split for model development and validation.

The training process of the Models of each of the three languages involved a 5-fold cross-validation strategy with a batch size of 32 training patterns. The models were trained for five epochs across each fold using the Adam optimizer, chosen for its efficiency in handling non-stationary objectives and providing adaptive learning rates and Categorical Crossentropy as the loss function.

\subsubsection{Task 2}

Task 2 required us to use transfer learning from other open datasets concerning hate speech and toxic text recognition in the Indic languages to create a classifier for detecting gender-based abusive content (label 1). In addition to the provided dataset, we used the  Multilingual Abusive Comment Detection (MACD) dataset \cite{NEURIPS2022_a7c4163b}. for Hindi and Tamil, and the MULTILATE \footnote{\url{https://github.com/advaithavetagiri/MULTILATE}}
 dataset for English as external open datasets. 

The MULTILATE Dataset is a large labeled dataset containing over 2.6 million sentences for detecting Hate and Abusive Content in social media. The dataset has been annotated into ‘\textit{Hate}’ and ‘\textit{Not-Hate}’ for a Binary Classification Task and into ‘\textit{sexist}’, ‘\textit{racist}’ and ‘\textit{neither}’ for a Multiclass Classification Task. The MACD dataset comprises around 150K textual sentences with 74K abusive and 77K non-abusive comments from five Indic languages - Hindi (Hi), Tamil (Ta), Telugu (Te), Malayalam (Ml) and Kannada (Kn), annotated as 0 (For abusive) and 1 (For non-abusive). We used the Hindi and the Tamil subsets of the MACD dataset, consisting of 33k sentences and 30k sentences, respectively.

In Task 2, for each language, after assigning the final levels to each sentence in the provided dataset, we concatenated the given dataset with the respective external datasets to create a final combined dataset. The datasets were then divided into an 80/20 train-test split for model development and validation.

Training processes similar to Tasks 1 \& 3 were employed with 5-fold cross-validation but with a batch size of 64 training patterns and seven epochs for each fold. The models were trained using the Adam Optimiser and Categorical Crossentropy as the loss function.

\section{Results}

\subsection{Evaluation}

To evaluate our Models' overall performance and efficiency, we adopted several metrics, particularly precision, recall, and the F1 score. The F1 score is an essential evaluation measure because it consolidates the performance of a classifying model in terms of all categories into one statistic by providing a balanced measure between the model’s precision and recall. Precision measures how accurately the model predicts positive outcomes. At the same time, recall tells us how well the model captures all the relevant instances, giving us insights into how broadly the models can predict. These metrics are immensely useful in tasks such as classification, where finding a middle ground between accuracy and completeness is crucial. 

The precision and recall for the positive class can be calculated as follows:

\begin{equation}
    Precision_{1} = \frac{TP}{TP + FP}
\end{equation}

\begin{equation}
    Recall_{1} = \frac{TP}{TP + FN}
\end{equation}

Similarly, the precision and recall for the negative class can be calculated as:

\begin{equation}
    Precision_{2} = \frac{TN}{TN + FN}
\end{equation}

\begin{equation}
    Recall_{2} = \frac{TN}{TN + FP}
\end{equation}

Macro-Average Precision and Recall for Multiclass Classification:

\begin{equation}
    Precision_c = \frac{TP_c}{TP_c + FP_c}
\end{equation}

\begin{equation}
    Recall_c = \frac{TP_c}{TP_c + FN_c}
\end{equation}

The Macro-Average Precision (MAP) and Recall (MAR) can then be calculated as:

\begin{equation}
    MAP-Binary = \frac{Precision_{1} + Precision_{2}}{2}
\end{equation}

\begin{equation}
    MAR-Binary = \frac{Recall_{1} + Recall_{2}}{2}
\end{equation}

\begin{equation}
    MAP-Multiclass = \frac{1}{C} \sum_{c=1}^{C} Precision_c
\end{equation}

\begin{equation}
    MAR-Multiclass = \frac{1}{C} \sum_{c=1}^{C} Recall_c
\end{equation}

Finally, the Macro F1 score can be computed using the formula:

\begin{equation}
    F1_{\text{macro}} = \frac{2 \times (MAP \times MAR)}{(MAP + MAR)}
\end{equation}
To measure the efficiency of our models, we have used the above metrics, the results of which have been mentioned in the next section.

\subsection{Training Results}

To analyze the training performance, we generated classification reports for the Models of each language across each task, emphasizing on the precision, recall, and F1 scores for each fold and calculated the average macro scores for each of them. 

\begin{table}[!h]
    \centering
    \begin{tabular}{ccccc} \hline
        \textbf{Task} & \textbf{Language} & \textbf{P} & \textbf{R} & \textbf{F1}\\ \hline
         & English & 0.79 & 0.80 & 0.79\\
        Task 1 & Hindi & 0.73 & 0.75 & 0.70\\
         & Tamil & 0.75 & 0.74 & 0.74\\ \hline
         & English & \textbf{0.84} & \textbf{0.84} & \textbf{0.84}\\
        Task 2 & Hindi & 0.79 & 0.79 & 0.78\\
         & Tamil & 0.83 & 0.83 & 0.83\\ \hline
         & English & 0.79 & 0.80 & 0.79\\
        Task 3 & Hindi & 0.73 & 0.75 & 0.70\\
         & Tamil & 0.75 & 0.75 & 0.74 \\
         \hline
    \end{tabular}
    \caption{The training results as Precision (P), Recall (R), and F1 Scores (F1) on the languages English, Hindi, and Tamil by the models CNN-BiLSTM using GloVe Embeddings for English and FastText Embeddings for Hindi and Tamil.}
    \label{tab:testing_results}
\end{table}
 \begin{center}
      \includegraphics[width=1\linewidth]{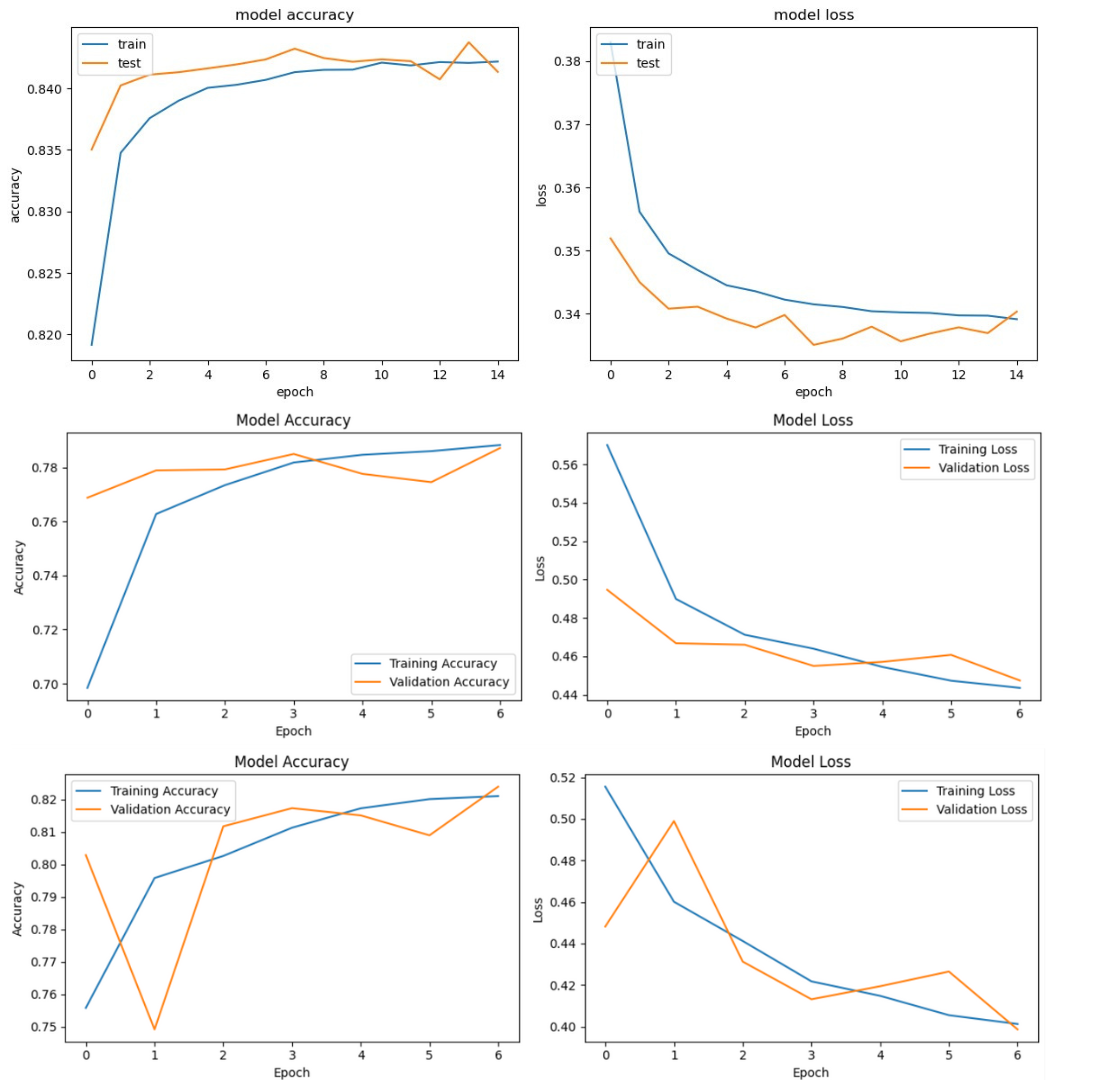}
      \captionof{figure}{Task 2 Accuracy and Loss of English (Top), Hindi (Middle) and Tamil(Bottom).}
     \label{fig:figlabel}
    \end{center}

The training performance metrics scores for the Models have been depicted in Table \ref{tab:testing_results}. The performance for all the models has been sufficiently high across all tasks. For all tasks 1, 2 \& 3, the precision, recall, and F1 score has been the highest for English, followed by Tamil and Hindi. The training scores in Task 2 for all the models are considerably higher than the other two tasks, primarily due to the availability of larger datasets for training. The highest F1- the English Model has achieved a score across all tasks in Task 2 at 84\%. 
For all the matrices to be measured with accordance with the task given, there have been instances which have taken the form of alias and as such there needs to be alterations that need to be made in the name 

We also generated the graphs representing the Model Accuracy as a function of the number of epochs across the training and the validation sets and the Model Loss as a function of the number of epochs across the training and the validation sets to check for overfitting and model improvement during the entire training duration of the Models, the plots for which have been demonstrated in Figure \ref{fig:figlabel}.

All the graphs show an increase in the validation accuracy and a decrease in the validation loss across all the epochs for a corresponding increase in the training accuracy and a decrease in the training loss, respectively, indicating the generation of a robust model with improved generalization and no overfitting.

\subsection{Testing Results}

We predicted the labels for test datasets provided for each task using the models for all three languages, compiled them into .csv files, and submitted them in the respective Kaggle competitions to obtain the test set evaluation scores for each task. The F1 score was used as the testing metric for the shared tasks, as it provides a balanced evaluation basis between precision and recall and is also known to give good results on imbalanced classification problems.

\begin{table}[!h]
    \centering
    \begin{tabular}{ccc}\hline
         & \textbf{Tasks} & \textbf{F1 Score}\\ \hline
         & Task 1 & 0.616\\
        CNLP-NITS-PP & Task 2 & 0.572\\
         & Task 3(Multi) &\textbf{ 0.616} \& \textbf{0.582} \\ \hline
         
    \end{tabular}
    \caption{Testing Results of the CNN-BiLSTM models on each task}
    \label{tab:my_label}
\end{table}

The scores for the Models across each of the three tasks have been summarized in Table \ref{tab:my_label}. In Task 1, we achieved the highest F1 score of 0.616. For Task 2, the F1-score for our Models came out to be 0.572, and for Task 3, our Models scored F1-scores of 0.616 for Label 1 and 0.582 for Label 3.

\subsection{Results Analysis}

Our study reveals promising performance outcomes for the various classification tasks under the shared tasks of ICON 2023. The model performance analysis across all three tasks and datasets reveals intriguing patterns in training and testing scenarios. 

From Table \ref{tab:testing_results}, we observe that the training set performance of the Models, across all the languages, for Task 2 is better than that of Task 1 by a considerable margin. These improvements in the performance scores for Task 2 can be attributed to the utilization of significantly larger volumes of training data from external sources, hinting at improved generalization capabilities of the models, as they benefit from a more diverse and extensive set of examples. In addition, the models seem to perform considerably well in the training set evaluation for Task 3.

Contrary to the patterns observed in the training set, when assessed on the test set, the models exhibited higher scores for Task 1 than Task 2, as evident from Table \ref{tab:my_label}. The reason for this change in trend cannot be explained directly based on our experimental findings. This is due to the collective evaluation of the test set performance across all languages in each task, making it difficult for us to verify the contributions of the models pertaining to each language. In contrast, the test set results for Task 3 conform to those of the training set results as expected.

\section{Conclution}
This paper presented our approach and results for the ICON2023 shared task on identifying gendered abuse in online content. Our ensemble models using CNN-BiLSTMs and contextual embeddings like FastText proved effective, achieving top ranks on the leaderboard across multiple languages. The models could capture nuanced abusive language through localized feature learning and sequence modelling. Our analysis showed the impact of factors like embedding techniques and input preprocessing. There is still difficulty in handling heavily code-switched languages - an area for future work. Through this shared task, we developed performant models for a crucial problem limiting online freedom of expression. The datasets and model code have been open-sourced to enable further research towards mitigating such gendered cyber harassment.

\section*{Acknowledgements}
 We appreciate the Department of Computer Science \& Engineering, National Institute of Technology Silchar, for allowing us to pursue our research and experimentation, the Center for Natural Language Processing (CNLP) \ and Artificial Intelligence (AI) laboratories' resources, and the research atmosphere.

\bibliography{main}



\end{document}